\documentclass{article} 
\usepackage[preprint]{colm2026_conference}

\usepackage{microtype}
\usepackage{hyperref}
\usepackage{url}
\usepackage{booktabs}
\usepackage[textsize=small]{todonotes}
\usepackage{amsmath}
\usepackage{amsfonts}
\usepackage{amssymb}
\usepackage{amsthm}
\usepackage{bm}
\usepackage{enumitem}
\usepackage{multirow}
\usepackage{array}
\usepackage{mathtools}
\usepackage{algorithm}
\usepackage{algorithmic}
\usepackage[normalem]{ulem}
\usepackage{cleveref}
\usepackage{wrapfig}
\usepackage{tcolorbox}
\usepackage{pifont}

\usepackage{caption}
\newsavebox{\synablationbox}

\usepackage{graphicx}
\usepackage{subcaption}

\usepackage{lineno}

\definecolor{darkblue}{rgb}{0, 0, 0.5}
\hypersetup{colorlinks=true, citecolor=darkblue, linkcolor=darkblue, urlcolor=darkblue}

\title{
OLIVIA: Online Learning via Inference-time Action Adaptation for Decision Making in LLM ReAct Agents
}

\author{
Sheldon Yu$^{1}$\thanks{These authors contributed equally to this work.}, Junda Wu$^{1*}$, Xintong Li$^{1}$, Nikki Lijing Kuang$^{1}$, Sizhe Zhou$^{2}$, \\
\textbf{Tong Yu$^{3}$, Jiawei Han$^{2}$, Jingbo Shang$^{1}$, Julian McAuley$^{1}$}
\\
$^{1}$UC San Diego \quad
$^{2}$University of Illinois at Urbana-Champaign \quad
$^{3}$Adobe Research \\
\texttt{\{ziy040,juw069,xil240,jshang,jmcauley\}@ucsd.edu} \\
\texttt{\{sizhez,hanj\}@illinois.edu} \quad
\texttt{tyu@adobe.com} \\
}

\theoremstyle{plain}
\newtheorem{assumption}{Assumption}[section]

\theoremstyle{remark}
\newtheorem{remark}[assumption]{Remark}

\begin{document}

\ifcolmsubmission
\linenumbers
\fi

\maketitle

\begin{abstract}
Large language model agents interleave reasoning, action selection, and observation to solve sequential decision-making tasks. 
In deployed settings where agents repeatedly handle related multi-step tasks, small action-selection errors can accumulate into wasted tool calls, latency, and reduced reliability. Despite this need for deployment-time improvement,
however, existing inference-time adaptation methods for LLM agents mainly rely on prompting or retrieval, which influence behavior indirectly through context manipulation. For ReAct-style agents, such approaches do not expose an explicit decision layer that can score candidate actions, represent uncertainty, or be updated online from action-level feedback. As a result, they provide limited support for trackable, fine-grained, and uncertainty-aware adaptation during deployment.
We propose \textsc{OLIVIA}, an inference-time action adaptation framework for ReAct-style agents. \textsc{OLIVIA} models the LLM’s final action-selection layer as a contextual linear bandit over candidate actions, with frozen hidden states as decision contexts. This choice is particularly suitable for deployment because it adapts behavior directly at the action-selection interface, preserves the underlying reasoning process, and provides explicit uncertainty estimates and lightweight online updates from action-level feedback. With upper-confidence-bound exploration, \textsc{OLIVIA} improves the policy sample-efficiently with minimal computational overhead.
We instantiate \textsc{OLIVIA} on four benchmarks and show that it consistently improves task performance over static ReAct and prompt-based inference-time baselines. 
Our results suggest that explicit online decision layers provide an effective alternative to purely prompt or retrieval-based adaptation for LLM agents during deployment.
\end{abstract}
\section{Introduction}
LLM agents increasingly operate as sequential decision makers in deployed settings such as tool-using assistants, API orchestration systems, and enterprise workflow agents. In these applications, the agent must interleave natural-language reasoning, action selection, and environment feedback through a sequence of thoughts, actions, and observations in order to complete multi-step tasks. ReAct-style prompting has emerged as a practical template for such agents because it exposes intermediate decisions, supports multi-step interaction, and can be deployed without parameter updates~\citep{wu2025docreact, nguyen2025guisurvey, xia2025sand}. However, despite this sequential structure, agent behavior is typically \emph{static} at inference time: the model follows a fixed prompt-guided policy, while feedback gathered during deployment is not used to systematically improve future decisions. In repeated or streaming deployments, this limitation can be costly. Agents may revisit similar decision states, repeat the same local mistakes, and waste interaction budget when an early error leads to an unproductive trajectory. A growing body of work seeks to improve inference-time behavior through prompt engineering \citep{yao2023react}, reflection \citep{shinn2023reflexion}, or memory augmentation \citep{majumder2023clin, wang2025dice, wang2025selfupdatable}. These approaches can change what the model generates, but they typically do not provide an explicit online optimization mechanism for sequential decision making~\citep{huang2025agenticrs}. In particular, they offer limited support for adapting action-selection behavior from observed feedback
over a stream of related tasks. As a result, improvements often remain indirect, heuristic, or tied to prompt design rather than to an explicit decision layer that can learn during deployment.

We study this problem through the lens of \emph{online adaptation for inference-time decision making}. At each decision step, an agent must choose an action based on the current task state and its evolving reasoning history, while feedback from previous interactions may be informative for future decisions in similar contexts~\citep{wu2022dynamicsrl}. This setting naturally calls for a lightweight online learner that can sit on top of a frozen LLM, use the model's internal state as context, and update from observed interaction feedback without requiring model fine-tuning.

To this end, we propose \textsc{OLIVIA}, an online adaptation framework for ReAct-style agents. \textsc{OLIVIA} adds a lightweight contextual decision layer on top of a frozen language model: at each step, it uses the model's hidden-state representation of the current task and reasoning history as context, selects the next action from a candidate set, and updates from step-level feedback to improve future decisions across a stream of tasks. This design is well matched to our setting: the agent repeatedly chooses from a discrete action set and observes feedback only for the selected action, making contextual bandits a natural model of partial-feedback online adaptation~\citep{wu2021clusterbandit, hu2025hiersucb}. Compared with policy-gradient or purely episodic updates, this yields a more direct and lower-variance learning signal, while unlike retrieval-augmented memory, it adapts behavior through an explicit decision layer rather than indirect context modification. In this paper, we instantiate \textsc{OLIVIA} on tool-use benchmarks, where decision context is trajectory-dependent: each action changes the subsequent reasoning history, so later hidden states, rewards, and updates are coupled across the interaction even though the backbone model remains fixed.

Our approach is lightweight and modular. It does not require retraining the base model, modifying the prompting interface, or storing large retrieval memories. Instead, it augments the frozen LLM with an online controller that adapts action preferences from experience, enabling the agent to improve over time as it encounters related tasks. We summarize our contributions as follows:
\begin{itemize}
    \item We formulate online adaptation for language model agents as an inference-time sequential decision-making problem, in which a lightweight decision layer uses LLM hidden-state representations to adapt action selection from step-level feedback.
    \item We develop \textsc{OLIVIA}, a contextual-bandit-based online decision layer that operates on top of a frozen LLM and supports efficient action scoring, uncertainty-aware exploration, and incremental updates during deployment.
    \item We instantiate \textsc{OLIVIA} on four tool-use benchmarks and show that it improves task performance and online adaptation behavior over static prompting baselines and a prompt-based inference-time adaptation baseline, while preserving the underlying language model.
\end{itemize}

\begin{figure}
    \centering
    \includegraphics[width=1.\linewidth]{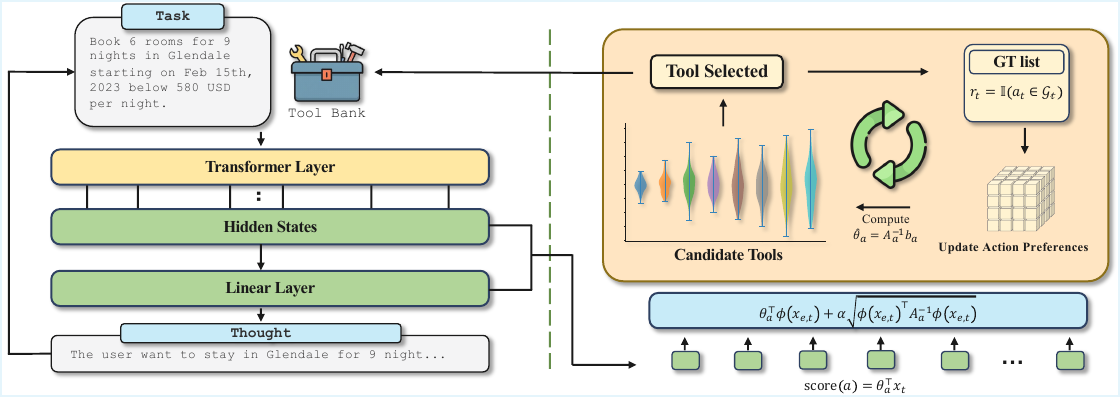}
    \caption{Overview of \textsc{OLIVIA}. \textbf{Left}: A frozen LLM backbone processes the task and trajectory prefix through its transformer layers, producing hidden states and a reasoning trace (\texttt{Thought}). \textbf{Right}: At each action-selection step, \textsc{OLIVIA} extracts the last-layer hidden state as the decision context and scores candidate tools using per-action UCB estimates.}
    \label{main_fig}
\end{figure}

\section{Related work}
\subsection{Inference-time adaptation.}
Recent work shows that LLM agents need not remain fixed after deployment. ReAct already
casts inference as a closed-loop process of reasoning, acting, and observing, creating a
natural interface for adaptation from interaction feedback \citep{yao2023react, wu2025docreact, nguyen2025guisurvey, lu2024uci}. Most
inference-time adaptation methods, however, operate indirectly through memory, prompting,
or lightweight control variables: verbal reflection and episodic memory
\citep{majumder2023clin, shinn2023reflexion, zhong2023memorybank}, dynamic in-context example selection \citep{wang2025dice}, in-context preference adaptation \citep{wu2025icrpo}, and parameter-internalized context \citep{wang2025selfupdatable}, reusable natural-language experience or
skill libraries \citep{zhao2024expel, wang2023voyager,xia2025sand}, retrospective prompt optimization
\citep{yao2024retroformer}, CoT debiasing and offline alignment \citep{wu2024decot, wu2025ocean}, and information-theoretic prompt tuning \citep{wu2023infoprompt}, and lightweight test-time adaptation under environment shift
\citep{chen2026testtime}, latent-state reasoning analysis \citep{wu2025ctrls, yu2025explainablecot}. Surveys also document the rapid expansion of inference-time adaptation across personalized, multimodal, and federated LLMs~\citep{zhang2024llmpersonalization, wu2024mllmsurvey, wu2024visualpromptsurvey, surana2026generatefiltercontrolreplay}. These works establish deployment-time adaptation as a viable
paradigm, but they generally do not expose an explicit online decision layer for sequential
action selection. Our work shares the same deployment-time perspective, but focuses on
updating a lightweight action-selection module online from step-level feedback.

\subsection{Online optimization of sequential decisions.}
A complementary line of work treats inference itself as an online optimization problem over
a sequence of intermediate decisions. In reasoning, this appears as explicit search or
planning over thought states \citep{yao2023tree, hao2023reasoning}; in agent settings, as
tree-search-based interaction and planning with environment feedback
\citep{zhou2024language, koh2026tree,mundada2026wsgrpo}; and in test-time compute scaling, as adaptive
allocation of inference resources such as branching depth or search breadth
\citep{inoue2025wider, liu2024sd,kveton2025aldpo}. Related uncertainty-aware methods also use posterior
sampling or contextual Thompson sampling to guide sequential choice under partial feedback
\citep{cai2025active, zhang2025contextual,xia2023deconfoundedcrs}. Together, these works motivate viewing
inference as online optimization over reasoning, acting, and compute allocation. Our method
is closest to this perspective: we study online adaptation for sequential LLM decision
making using frozen-model contextual representations and lightweight online updates from
observed step-level feedback. Closely related online-RL formulations adapt action policies under simulator-to-deployment dynamics shift~\citep{wu2022dynamicsrl}, while pluralistic off-policy evaluation extends this view to multi-stakeholder alignment~\citep{huang2025pluralistic, xie2025survey}.

\section{Preliminaries}
\label{sec:prelim}
\subsection{Contextual Bandits}
\label{sec:prelim_cb}

A contextual bandit models repeated decision making under partial feedback
\citep{Li_2010, ban2023neural}. At round $n = 1,2,\dots,N$,
the learner observes a context $x_n \in \mathcal{X}$, chooses an action
$a_n \in \mathcal{A}(x_n)$, and receives a scalar reward $r_n$ for that action alone;
the rewards of unchosen actions are not observed. An optimal action satisfies
\[
a_n^* \in \arg\max_{a \in \mathcal{A}(x_n)} \mathbb{E}[r_n \mid x_n, a].
\]
The learner's goal is to find an action-selection rule that achieves high cumulative
reward across rounds while balancing \emph{exploration} of uncertain actions against
\emph{exploitation} of actions that already appear promising. When the expected reward
is modeled as a parametric function of contextual features, the learner can generalize
across related decision states and employ principled exploration strategies such as
upper confidence bounds \citep{ban2023neural}.

\subsection{ReAct-Style Sequential Agents}
\label{sec:prelim_agent}

We consider an LLM agent that interacts with an environment over multiple decision
steps following the ReAct paradigm \citep{yao2023react, zhou2024language, koh2026tree}.
In episode $e$, the agent receives a task input $q_e$ and a candidate action set
$\mathcal{A}_e$. Let $h_{e,t-1}$ denote the interaction history before step $t$,
comprising prior reasoning traces, selected actions, and environment observations.
At step $t$ the agent selects an action $a_{e,t} \in \mathcal{A}_{e,t}$, receives an
observation $o_{e,t}$, and updates the history:
\[
h_{e,t} = h_{e,t-1} \oplus (a_{e,t}, o_{e,t}),
\]
where $\oplus$ denotes concatenation. The episode terminates when the agent issues a
stop action or reaches a step budget. Because the history grows with each step, every
action decision is conditioned on a distinct, trajectory-dependent context
$x_{e,t} = (q_e, h_{e,t-1})$.

\section{Problem Formulation}
\label{sec:problem}

We study \emph{online inference-time action adaptation} for sequential LLM
agents. Our goal is to enable a frozen language model to improve its action
choices over a stream of related episodes by updating only a lightweight
decision layer from interaction feedback, without fine-tuning the backbone.
This section develops the conceptual foundation in three steps: we first show
that ReAct-style action selection naturally maps to a contextual bandit
problem, then identify a linear decision interface inside the frozen LLM that
can serve as the bandit's scoring mechanism, and finally argue that this
structure enables principled online adaptation with exploration.

\subsection{ReAct Action Selection as a Contextual Bandit}
\label{sec:problem_bandit}

Consider a stream of ReAct-style episodes $e = 1,2,\dots,N$. Using the
notation from Section~\ref{sec:prelim_agent}, at step $t$ of episode $e$ the
agent faces a decision context $x_{e,t} = (q_e, h_{e,t-1})$ and must choose
the next action from $\mathcal{A}_{e,t}^{\mathrm{valid}} \subseteq
\mathcal{A}_e$, the subset of actions still available at that step. After
acting, the agent observes feedback only for the selected action---the rewards
of alternative actions remain unknown.

This per-step structure directly mirrors the contextual bandit setting of
Section~\ref{sec:prelim_cb}: a context-dependent choice from a discrete
action set under partial feedback. The key additional observation is that,
although full trajectories differ across tasks, \emph{similar local decision
states recur} across a stream of related episodes. When the same or a closely
related action-selection problem appears again, feedback collected from earlier
decisions becomes informative. This recurrence creates an opportunity for
deployment-time learning that purely static prompting cannot exploit.
We therefore cast each action-selection step in the ReAct loop as a round of
an online contextual decision problem. The remainder of this section develops
the representation and adaptation mechanism that make this casting concrete.

\subsection{The LLM's Action-Selection Interface as a Linear Bandit}
\label{sec:problem_llm}

The bandit perspective above requires a contextual representation of each
decision state and a scoring rule over candidate actions. We now show that the
frozen LLM already provides both.

At the token position immediately preceding the next action decision, the
model's last-layer hidden state
\begin{equation}
\label{eq:hidden_state}
u_{e,t} = h_{\mathrm{LLM}}(q_e, h_{e,t-1}) \in \mathbb{R}^d
\end{equation}
encodes the task semantics and the full trajectory so far. Intuitively, this
vector is the model's internal answer to the local question: \emph{what action
should come next?}

To produce a next-token distribution, the frozen model scores each vocabulary
token through its output projection layer. When the output vocabulary is
restricted to the candidate action set $\mathcal{A}_{e,t}^{\mathrm{valid}}$,
this projection reduces to a per-action linear score
\begin{equation}
\label{eq:frozen_score}
s_0(u_{e,t}, a) = w_a^\top u_{e,t},
\end{equation}
where $w_a \in \mathbb{R}^d$ is the frozen parameter row associated with
action $a$---concretely, the corresponding row of the language-model head or
an embedding derived from the same model.

This decomposition reveals that the frozen LLM's action-selection mechanism is
already a \emph{contextual linear scorer}: the hidden state serves as the
context vector and the per-action parameters serve as arm-specific weights, exactly the
structure assumed by linear contextual bandits. We emphasize that we are not
claiming the full LLM is a bandit; rather, we isolate its
\emph{action-selection interface} as a linear scoring layer that can be
separated from the underlying reasoning dynamics.

This observation is the conceptual basis of our method. Because the frozen
scoring layer $\{w_a\}$ is fixed, it cannot adapt to deployment feedback. But
the same linear structure admits a \emph{learnable} replacement: a per-action
parameter $\theta_a$ that starts from the LLM's prior and is updated online as
the agent accumulates interaction experience. The next subsection describes the
adaptation mechanism that this structure enables.

\subsection{Online Adaptation and Exploration}
\label{sec:problem_online}

Replacing the frozen scoring weights with learnable parameters creates an
online learning problem: the agent must estimate which actions yield high
reward in each context from a growing stream of partial-feedback interactions.
A critical challenge is that feedback is observed \emph{only for the selected
action}. If the learner always picks the action that currently looks best, it
may never discover that an under-explored action is superior in a given
context.

Existing inference-time adaptation methods do not address this challenge
directly. Prompt-based approaches such as reflection \citep{shinn2023reflexion}
and memory augmentation \citep{majumder2023clin} influence behavior indirectly
through context manipulation, but they have no mechanism to systematically
explore uncertain actions or to quantify how confident the agent is about each
candidate.

The linear bandit structure identified in Section~\ref{sec:problem_llm}
provides a principled solution. Because the learner maintains an explicit
parametric model of action quality, it can estimate not only the expected reward
of each action but also the \emph{uncertainty} of that estimate. An optimistic
selection rule can then add an exploration bonus proportional to the current
uncertainty, favoring actions whose reward is still poorly characterized. As
more feedback is collected, the uncertainty shrinks and the learner
concentrates on high-reward actions---balancing exploration and exploitation
automatically.
\section{Methodology}
\label{sec:method}

This section presents \textsc{OLIVIA}, the concrete instantiation of the
framework outlined in Section~\ref{sec:problem}. \textsc{OLIVIA} operates
entirely at the action-selection interface of a frozen LLM: the model's
reasoning process generates a trajectory-dependent hidden state, and a
lightweight online decision layer selects the next action from the valid
candidate set based on that hidden state. All backbone parameters, prompts, and
reasoning generation remain fixed throughout deployment; only the decision
layer is updated from online feedback.

\subsection{Hidden-State Decision Layer}
\label{sec:method_layer}

At each decision step $t$ of episode $e$, the frozen LLM processes the current
task $q_e$ and trajectory prefix $h_{e,t-1}$. We extract the last-layer hidden
state at the token position immediately preceding the next action token:
\begin{equation}
\label{eq:context}
u_{e,t} = h_{\mathrm{LLM}}(q_e, h_{e,t-1}) \in \mathbb{R}^d.
\end{equation}
This vector serves as the contextual input to the online controller. It
encodes both the original task semantics and the evolving reasoning history, so
two decision points that share similar task structure and trajectory prefixes
produce similar context vectors, enabling generalization across episodes.

For each candidate action $a$, \textsc{OLIVIA} maintains per-action sufficient
statistics
\begin{equation}
\label{eq:stats}
A_a \in \mathbb{R}^{d \times d}, \qquad
b_a \in \mathbb{R}^d, \qquad
\hat{\theta}_a = A_a^{-1} b_a,
\end{equation}
where $\hat{\theta}_a$ is the current estimate of the action's reward
parameter. The base predicted reward for action $a$ in context $u_{e,t}$ is
\begin{equation}
\label{eq:base_score}
s_{e,t}(a) = \hat{\theta}_a^\top u_{e,t}.
\end{equation}

\paragraph{LLM-aligned initialization.}
Rather than starting from a random policy, we warm-start the controller from
the frozen model's own prior. We initialize
\begin{equation}
\label{eq:init}
A_a^{(0)} = I, \qquad b_a^{(0)} = \mathrm{embed}(a),
\end{equation}
where $\mathrm{embed}(a) \in \mathbb{R}^d$ is the mean input embedding of the
action's name tokens under the same LLM. Because $A_a^{(0)} = I$, the initial
parameter satisfies $\hat{\theta}_a^{(0)} = \mathrm{embed}(a)$, so the
controller's starting scores are aligned with the frozen model's
representation of each action. This initialization is especially valuable early
in the stream before substantial feedback has been collected, as it ensures the
controller inherits the semantic priors already encoded in the language model.

\subsection{UCB-Based Exploration}
\label{sec:method_ucb}

The partial-feedback nature of the problem (Section~\ref{sec:problem_online})
demands a selection rule that actively explores uncertain actions. We adopt the
LinUCB strategy \citep{Li_2010}, which augments the predicted reward with an
exploration bonus that reflects the learner's uncertainty about each action in
the current context.

Given context $u_{e,t}$, each valid action $a \in
\mathcal{A}_{e,t}^{\mathrm{valid}}$ is scored by
\begin{equation}
\label{eq:ucb}
\mathrm{UCB}_{e,t}(a)
= \hat{\theta}_a^\top u_{e,t}
+ \alpha \sqrt{u_{e,t}^\top A_a^{-1} u_{e,t}},
\end{equation}
where $\alpha > 0$ is the exploration coefficient. The first term is the
current reward estimate; the second term is the exploration bonus,
proportional to the width of the confidence interval for action $a$ in the
direction of the current context. The selected action is
\begin{equation}
\label{eq:select}
a_{e,t}
\in
\arg\max_{a \in \mathcal{A}_{e,t}^{\mathrm{valid}}}
\mathrm{UCB}_{e,t}(a).
\end{equation}

The exploration bonus has two desirable properties. First, it is
\emph{context-dependent}: the bonus for a given action varies with the
decision state, so the controller explores actions that are uncertain in the
current context rather than exploring uniformly at random. Second, it
\emph{shrinks automatically} as data accumulates: each observation tightens
the confidence ellipsoid, reducing the bonus for well-characterized actions and
concentrating selection on those that remain uncertain.
These properties distinguish UCB from simpler alternatives.
$\varepsilon$-greedy exploration selects randomly and ignores the structure of
the context space; Thompson sampling introduces stochastic noise that can be
harder to diagnose in deployment; prompt-based adaptation methods have no
explicit exploration mechanism at all.

The valid action set $\mathcal{A}_{e,t}^{\mathrm{valid}} \subseteq
\mathcal{A}_e$ may shrink during an episode: once an action has been selected
the maximum allowed number of times, it is removed from the candidate set.
This constraint prevents degenerate repetition while still allowing the
controller to select the same tool more than once when warranted by the
ground-truth trajectory.

The coefficient $\alpha$ controls exploration aggressiveness. Smaller values
place more weight on the current reward estimate and the LLM-aligned prior;
larger values encourage broader exploration early in the stream. We study the
sensitivity to $\alpha$ empirically in Section~\ref{sec:ablation}.

\subsection{Step-Level Feedback and Online Update}
\label{sec:method_update}

\begin{figure}[ht]
    \centering
    \small
    \includegraphics[width=1.\linewidth]{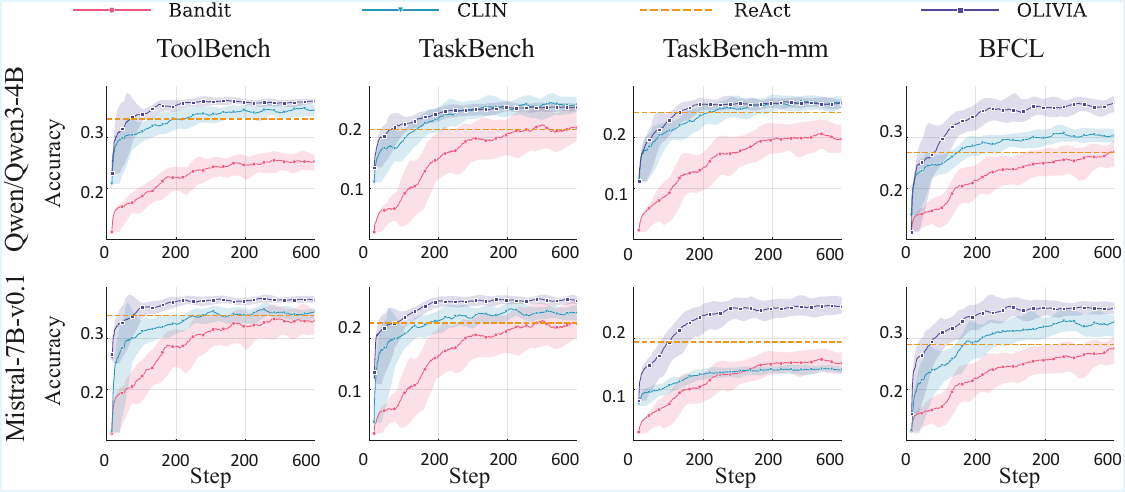}
    \caption{Running-average F1 over the episode stream on the four benchmarks. Static baselines (CoT, BM25, ReAct) remain flat; online methods improve over time. \textsc{OLIVIA} converges faster and reaches higher final F1 than \textsc{CLIN}.}
    \label{fig:learning_curves}
\end{figure}

\paragraph{Reward signal.}
In our tool-use instantiation, each episode $e$ is associated with a
ground-truth tool multiset $G_e$. Let $G_{e,t}^{\mathrm{rem}}$ denote the
unmatched portion of that multiset before step $t$. After selecting action
$a_{e,t}$, the agent receives immediate binary feedback:
\begin{equation}
\label{eq:reward}
r_{e,t} = \mathbf{1}[a_{e,t} \in G_{e,t}^{\mathrm{rem}}].
\end{equation}
If $r_{e,t} = 1$, one occurrence of $a_{e,t}$ is removed from
$G_{e,t}^{\mathrm{rem}}$ so that each ground-truth tool can be matched at most
once. This provides a step-level learning signal that directly reflects
action-selection quality, in contrast to episode-level feedback that conflates
all decisions within a trajectory.

\paragraph{Online update.}
After observing reward $r_{e,t}$, only the statistics of the selected action
are updated via a rank-one rule:
\begin{equation}
\label{eq:update_A}
A_{a_{e,t}} \leftarrow A_{a_{e,t}} + u_{e,t} u_{e,t}^\top,
\qquad
b_{a_{e,t}} \leftarrow b_{a_{e,t}} + r_{e,t}\, u_{e,t}.
\end{equation}
The parameter estimate is then $\hat{\theta}_{a_{e,t}} = A_{a_{e,t}}^{-1}
b_{a_{e,t}}$. In practice, we maintain the inverse $A_a^{-1}$ incrementally
using the Sherman--Morrison formula, so each update requires $O(d^2)$ time and
avoids re-inverting the full covariance matrix.

The episode terminates when the agent emits a stop action, no valid action
remains, or the per-episode step budget is exhausted. Across the full episode
stream, the cumulative reward is the objective we seek to maximize through online adaptation.

\paragraph{Design summary.}
The three components above---hidden-state context extraction, UCB-based
exploration, and rank-one online updates---together constitute the
\textsc{OLIVIA} decision layer. All backbone LLM parameters, the prompt
template, and the reasoning generator remain frozen throughout deployment. The
only learned quantities are the per-action statistics
$\{A_a, b_a\}$, which are lightweight and updated incrementally from
step-level feedback.

\section{Experiments}
\label{sec:exp}

\subsection{Experimental Setup}
\label{sec:exp_setup}

\textbf{Benchmarks.}
We evaluate on four tool-use benchmarks: \textsc{ToolBench} \citep{xu2023tool}, \textsc{TaskBench}, \textsc{TaskBench-MM} \citep{shen2024task}, and \textsc{BFCL} \citep{patil2023bfcl}. These benchmarks span complementary tool-use settings, varying in task domain, modality, and candidate-set size. Each instance provides a natural-language task, a candidate tool set, and one or more reference tool-call trajectories. Following our formulation, we convert each reference completion into a ground-truth tool multiset and evaluate all methods as sequential tool-selection policies under the same candidate tool set.

\textbf{Backbone models.}
We use \texttt{Qwen/Qwen3-4B} \citep{yang2025qwen3} and \texttt{mistralai/Mistral-7B-v0.1} \citep{jiang2023mistral7b} as base language models, chosen as representative open-weight backbones that support ReAct-style tool use while remaining practical for repeated online evaluation. We keep the backbone, system prompt, and few-shot demonstrations fixed across all methods so that performance differences can be attributed to the decision policy alone.

\textbf{Baselines.}
We compare against both non-learning and inference-time adaptation baselines. \textbf{CoT} \citep{wei2023chain} uses chain-of-thought prompting for free-form reasoning and tool selection without online adaptation. \textbf{BM25} \citep{10.1561/1500000019} ranks candidate tools by lexical similarity to the task description. \textbf{ReAct} \citep{yao2023react} performs iterative reasoning and tool selection but does not adapt across episodes. \textbf{CLIN} \citep{majumder2023clin} is a memory-based continual-adaptation method that updates a persistent textual memory after each episode to improve future decisions.

\textbf{Online protocol.}
Each dataset is treated as a stream of episodes. For every seed, we randomly shuffle the episode order and run each method sequentially over that stream, reporting mean and standard deviation over three random seeds. All methods use the same per-episode step budget $B = |G| + s$, where $|G|$ is the ground-truth tool-multiset size and $s$ is a small fixed allowance for recovery from early mistakes, and each tool is capped at $m$ selections per episode to prevent degenerate repetition.

\begin{table*}[t]
    \small
    \centering

    \label{tab:main_f1}
    \begin{tabular}{ll|cccc}
    \toprule
    & & \multicolumn{4}{c}{Dataset} \\
    \cmidrule(lr){3-6}
    Model & Method & ToolBench & TaskBench & TaskBench-MM & BFCL \\
    \midrule
    \multirow{6}{*}{\texttt{qwen}} 
      & CoT            & 0.211 & 0.030 & 0.031 & 0.082 \\
      & BM25           & 0.353 & 0.116 & 0.127 & 0.181 \\
      & ReAct          & 0.520 & 0.174 & 0.162 & 0.272 \\
      \cmidrule{2-6}
      & Bandit         & 0.348 & 0.155 & 0.126 & 0.242 \\
      & CLIN           & \underline{0.553} & \textbf{0.211} & \underline{0.170} & \underline{0.307} \\
      & OLIVIA (Ours)  & \textbf{0.585} & \underline{0.205} & \textbf{0.178} & \textbf{0.366} \\
    \midrule
    \multirow{6}{*}{\texttt{mistral}} 
      & CoT            & 0.142 & 0.036 & 0.064 & 0.107 \\
      & BM25           & 0.353 & 0.116 & 0.127 & 0.181 \\
      & ReAct          & 0.386 & 0.178 & \underline{0.146} & 0.281 \\
      \cmidrule{2-6}
      & Bandit         & 0.348 & 0.155 & 0.126 & 0.242 \\
      & CLIN           & \underline{0.394} & \underline{0.193} & 0.127 & \underline{0.324} \\
      & OLIVIA (Ours)  & \textbf{0.420} & \textbf{0.208} & \textbf{0.207} & \textbf{0.352} \\
    \bottomrule
    \end{tabular}
      \caption{Main results (F1) across four benchmarks and two backbone models. First three baselines with fixed policies; following methods below are \emph{online learning} methods that adapt from deployment feedback.}
\end{table*}

\textbf{Evaluation Metrics.}
We evaluate tool-selection quality using multiset-matching precision, recall, and F1 \citep{patil2023bfcl}, reporting macro-averaged scores over episodes with F1 as the primary metric. For the synthetic experiments described in Section~\ref{sec:ablation}, we additionally report cumulative regret and parameter estimation error $\|\hat{\theta}_t - \theta^\star\|_2^2$.

\subsection{Results}
\label{sec:main_results}

\paragraph{Learning curves.}

Figure~\ref{fig:learning_curves} plots running-average F1 over the episode stream. Static baselines (CoT, BM25, ReAct) produce flat reference lines since their policies are fixed. Among the online methods, \textsc{CLIN} improves gradually through episodic memory updates, but \textsc{OLIVIA} improves faster and converges to a higher F1 across all four benchmarks. This demonstrates the core deployment-time claim: the lightweight decision layer converts sparse step-level feedback into steadily better action selection, without modifying the backbone model.

\subsection{Ablation Studies}
\label{sec:ablation}

\paragraph{Synthetic validation.}

\begin{figure}[t]
    \centering
    \includegraphics[width=\linewidth]{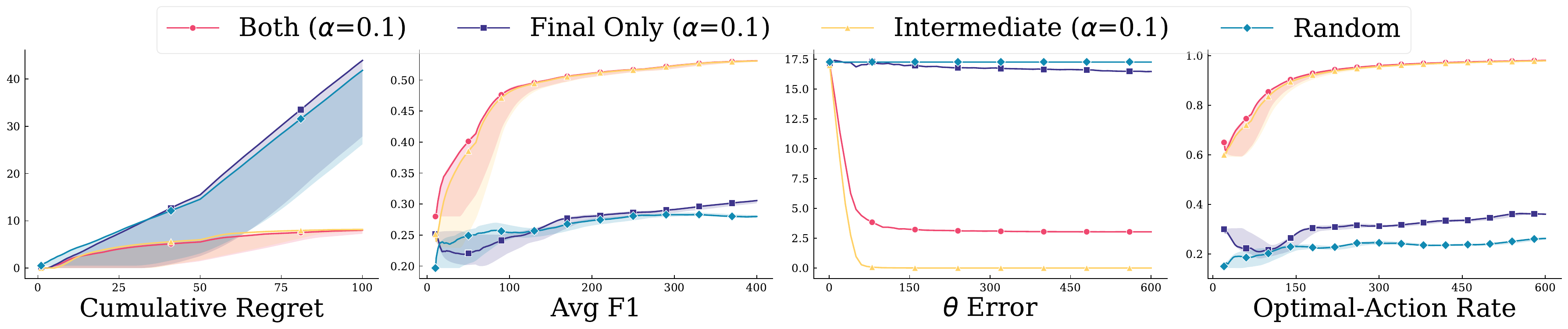}
    \caption{Synthetic experiments. Left: cumulative regret over rounds. Right: parameter estimation error $\|\hat{\theta}_t - \theta^\star\|_2^2$. \textsc{OLIVIA} converges faster and recovers the true reward parameter more accurately.}
    \label{fig:syn_results}
\end{figure}

To validate the online controller in a setting where the ground-truth reward structure is known, we construct synthetic contextual decision streams with linear rewards governed by an unknown $\theta^\star$ and partial feedback. Figure~\ref{fig:syn_results} shows that \textsc{OLIVIA} achieves lower cumulative regret and parameter estimation error than weaker online baselines, confirming that the controller recovers the true decision structure rather than improving by chance.

\paragraph{Effect of the exploration coefficient.}
We vary the LinUCB exploration coefficient $\alpha$ while keeping the rest of the method fixed. Table~\ref{tab:reward_ablation} shows that \textsc{OLIVIA} is robust across a reasonable range of $\alpha$ values but performs best in a moderate regime. When $\alpha$ is too small, the controller under-explores and is slower to correct early mistakes; when $\alpha$ is too large, excessive optimism leads to unnecessary exploration and weaker final performance.

\paragraph{Effect of reward design.}

\begin{wraptable}{r}{0.52\textwidth}
\vspace{-8pt}
\centering
\renewcommand{\arraystretch}{1.2}
\setlength{\tabcolsep}{0.9pt}
\caption{Reward-design ablation on synthetic streams.}
\label{tab:reward_ablation}
\begin{tabular}{
c
@{\hspace{3pt}}c@{\hspace{1pt}}c
@{\hspace{10pt}}cc
@{\hspace{10pt}}c
@{\hspace{10pt}}cc
}
\noalign{\hrule height 1.1pt}
\textbf{Reward}
& \multicolumn{1}{c}{\vline}
& \multicolumn{1}{c}{\vline}
& \multicolumn{2}{c}{\textbf{Performance}}
& \multicolumn{1}{c}{\vline}
& \multicolumn{2}{c}{\textbf{Convergence}} \\
& \multicolumn{1}{c}{\vline}
& \multicolumn{1}{c}{\vline}
& \shortstack{\textbf{Avg}\\\textbf{F1} ($\uparrow$)}
& \shortstack{\textbf{Opt.\ Rate}\\\textbf{(\%)} ($\uparrow$)}
& \multicolumn{1}{c}{\vline}
& \shortstack{\textbf{Regret}\\\textbf{@100} ($\downarrow$)}
& \shortstack{$\boldsymbol{\|\theta{-}\theta^*\|^2}$\\($\downarrow$)} \\
\noalign{\hrule height 0.8pt}
Final
& \multicolumn{1}{c}{\vline}
& \multicolumn{1}{c}{\vline}
& 0.323 & 36.2
& \multicolumn{1}{c}{\vline}
& 56.3 & 16.48 \\
Step
& \multicolumn{1}{c}{\vline}
& \multicolumn{1}{c}{\vline}
& 0.536 & 98.0
& \multicolumn{1}{c}{\vline}
& 8.2 & \textbf{0.00} \\
\noalign{\hrule height 0.8pt}
Both
& \multicolumn{1}{c}{\vline}
& \multicolumn{1}{c}{\vline}
& \textbf{0.536} & \textbf{98.2}
& \multicolumn{1}{c}{\vline}
& \textbf{8.3} & 3.04 \\
\noalign{\hrule height 1.1pt}
\end{tabular}
\vspace{-10pt}
\end{wraptable}

We compare three reward configurations: \emph{step-level only}, \emph{final only}, and \emph{both}. Table~\ref{tab:reward_ablation} shows that step-level feedback is substantially more effective than final-only reward, confirming that local credit assignment is important in multi-step tool selection. Combining both signals yields the strongest overall performance, suggesting that dense local feedback and episode-level evaluation provide complementary training signals.
\section{Conclusion} We presented \textsc{OLIVIA}, an inference-time action adaptation framework for ReAct-style agents. By placing an explicit online decision layer at the action-selection interface of a frozen language model, \textsc{OLIVIA} adapts behavior directly from action-level feedback while preserving the underlying reasoning process. Instantiated as a contextual linear bandit over hidden decision states, the method provides lightweight online updates, explicit uncertainty estimates, and sample-efficient exploration through UCB. Across four benchmarks, \textsc{OLIVIA} consistently improves task performance over static ReAct and prompt-based inference-time baselines, showing that deployment-time adaptation can be achieved without model fine-tuning. More broadly, our results suggest that explicit online decision layers are a promising alternative to purely prompt- or retrieval-based adaptation for sequential decision making in LLM agents.

\textbf{LLM usage:} AI writing tools were used to assist with drafting and the icons in Figure\ref{main_fig} are generated by AI drawing tool.

\bibliography{colm2026_conference}

@misc{yao2023react,
      title={ReAct: Synergizing Reasoning and Acting in Language Models}, 
      author={Shunyu Yao and Jeffrey Zhao and Dian Yu and Nan Du and Izhak Shafran and Karthik Narasimhan and Yuan Cao},
      year={2023},
      eprint={2210.03629},
      archivePrefix={arXiv},
      primaryClass={cs.CL},
      url={https://arxiv.org/abs/2210.03629}, 
}

@misc{shinn2023reflexion,
      title={Reflexion: Language Agents with Verbal Reinforcement Learning}, 
      author={Noah Shinn and Federico Cassano and Edward Berman and Ashwin Gopinath and Karthik Narasimhan and Shunyu Yao},
      year={2023},
      eprint={2303.11366},
      archivePrefix={arXiv},
      primaryClass={cs.AI},
      url={https://arxiv.org/abs/2303.11366}, 
}

@misc{zhao2024expel,
      title={ExpeL: LLM Agents Are Experiential Learners}, 
      author={Andrew Zhao and Daniel Huang and Quentin Xu and Matthieu Lin and Yong-Jin Liu and Gao Huang},
      year={2024},
      eprint={2308.10144},
      archivePrefix={arXiv},
      primaryClass={cs.LG},
      url={https://arxiv.org/abs/2308.10144}, 
}

@misc{wang2023voyager,
      title={Voyager: An Open-Ended Embodied Agent with Large Language Models}, 
      author={Guanzhi Wang and Yuqi Xie and Yunfan Jiang and Ajay Mandlekar and Chaowei Xiao and Yuke Zhu and Linxi Fan and Anima Anandkumar},
      year={2023},
      eprint={2305.16291},
      archivePrefix={arXiv},
      primaryClass={cs.AI},
      url={https://arxiv.org/abs/2305.16291}, 
}

@misc{zhong2023memorybank,
      title={MemoryBank: Enhancing Large Language Models with Long-Term Memory}, 
      author={Wanjun Zhong and Lianghong Guo and Qiqi Gao and He Ye and Yanlin Wang},
      year={2023},
      eprint={2305.10250},
      archivePrefix={arXiv},
      primaryClass={cs.CL},
      url={https://arxiv.org/abs/2305.10250}, 
}

@misc{yao2024retroformer,
      title={Retroformer: Retrospective Large Language Agents with Policy Gradient Optimization}, 
      author={Weiran Yao and Shelby Heinecke and Juan Carlos Niebles and Zhiwei Liu and Yihao Feng and Le Xue and Rithesh Murthy and Zeyuan Chen and Jianguo Zhang and Devansh Arpit and Ran Xu and Phil Mui and Huan Wang and Caiming Xiong and Silvio Savarese},
      year={2024},
      eprint={2308.02151},
      archivePrefix={arXiv},
      primaryClass={cs.CL},
      url={https://arxiv.org/abs/2308.02151}, 
}

@misc{chen2026testtime,
      title={Test-Time Adaptation for LLM Agents via Environment Interaction}, 
      author={Arthur Chen and Zuxin Liu and Jianguo Zhang and Akshara Prabhakar and Zhiwei Liu and Shelby Heinecke and Silvio Savarese and Victor Zhong and Caiming Xiong},
      year={2026},
      eprint={2511.04847},
      archivePrefix={arXiv},
      primaryClass={cs.LG},
      url={https://arxiv.org/abs/2511.04847}, 
}

@misc{yao2023tree,
      title={Tree of Thoughts: Deliberate Problem Solving with Large Language Models}, 
      author={Shunyu Yao and Dian Yu and Jeffrey Zhao and Izhak Shafran and Thomas L. Griffiths and Yuan Cao and Karthik Narasimhan},
      year={2023},
      eprint={2305.10601},
      archivePrefix={arXiv},
      primaryClass={cs.CL},
      url={https://arxiv.org/abs/2305.10601}, 
}

@misc{hao2023reasoning,
      title={Reasoning with Language Model is Planning with World Model}, 
      author={Shibo Hao and Yi Gu and Haodi Ma and Joshua Jiahua Hong and Zhen Wang and Daisy Zhe Wang and Zhiting Hu},
      year={2023},
      eprint={2305.14992},
      archivePrefix={arXiv},
      primaryClass={cs.CL},
      url={https://arxiv.org/abs/2305.14992}, 
}

@misc{zhou2024language,
      title={Language Agent Tree Search Unifies Reasoning Acting and Planning in Language Models}, 
      author={Andy Zhou and Kai Yan and Michal Shlapentokh-Rothman and Haohan Wang and Yu-Xiong Wang},
      year={2024},
      eprint={2310.04406},
      archivePrefix={arXiv},
      primaryClass={cs.AI},
      url={https://arxiv.org/abs/2310.04406}, 
}

@misc{koh2026tree,
      title={Tree Search for Language Model Agents}, 
      author={Jing Yu Koh and Stephen McAleer and Daniel Fried and Ruslan Salakhutdinov},
      year={2026},
      eprint={2407.01476},
      archivePrefix={arXiv},
      primaryClass={cs.AI},
      url={https://arxiv.org/abs/2407.01476}, 
}

@misc{liu2024sd,
      title={Speculative Decoding via Early-exiting for Faster LLM Inference with Thompson Sampling Control Mechanism}, 
      author={Jiahao Liu and Qifan Wang and Jingang Wang and Xunliang Cai},
      year={2024},
      eprint={2406.03853},
      archivePrefix={arXiv},
      primaryClass={cs.CL},
      url={https://arxiv.org/abs/2406.03853}, 
}

@misc{inoue2025wider,
      title={Wider or Deeper? Scaling LLM Inference-Time Compute with Adaptive Branching Tree Search}, 
      author={Yuichi Inoue and Kou Misaki and Yuki Imajuku and So Kuroki and Taishi Nakamura and Takuya Akiba},
      year={2025},
      eprint={2503.04412},
      archivePrefix={arXiv},
      primaryClass={cs.AI},
      url={https://arxiv.org/abs/2503.04412}, 
}

@misc{cai2025active,
      title={Active Exploration via Autoregressive Generation of Missing Data}, 
      author={Tiffany Tianhui Cai and Hongseok Namkoong and Daniel Russo and Kelly W Zhang},
      year={2025},
      eprint={2405.19466},
      archivePrefix={arXiv},
      primaryClass={cs.LG},
      url={https://arxiv.org/abs/2405.19466}, 
}

@misc{zhang2025contextual,
      title={Contextual Thompson Sampling via Generation of Missing Data}, 
      author={Kelly W. Zhang and Tiffany Tianhui Cai and Hongseok Namkoong and Daniel Russo},
      year={2025},
      eprint={2502.07064},
      archivePrefix={arXiv},
      primaryClass={cs.LG},
      url={https://arxiv.org/abs/2502.07064}, 
}

@misc{majumder2023clin,
      title={CLIN: A Continually Learning Language Agent for Rapid Task Adaptation and Generalization}, 
      author={Bodhisattwa Prasad Majumder and Bhavana Dalvi Mishra and Peter Jansen and Oyvind Tafjord and Niket Tandon and Li Zhang and Chris Callison-Burch and Peter Clark},
      year={2023},
      eprint={2310.10134},
      archivePrefix={arXiv},
      primaryClass={cs.CL},
      url={https://arxiv.org/abs/2310.10134}, 
}

@inproceedings{Li_2010, series={WWW ’10},
   title={A contextual-bandit approach to personalized news article recommendation},
   url={http://dx.doi.org/10.1145/1772690.1772758},
   DOI={10.1145/1772690.1772758},
   booktitle={Proceedings of the 19th international conference on World wide web},
   publisher={ACM},
   author={Li, Lihong and Chu, Wei and Langford, John and Schapire, Robert E.},
   year={2010},
   month=apr, pages={661–670},
   collection={WWW ’10} }

@misc{ban2023neural,
      title={Neural Contextual Bandits for Personalized Recommendation}, 
      author={Yikun Ban and Yunzhe Qi and Jingrui He},
      year={2023},
      eprint={2312.14037},
      archivePrefix={arXiv},
      primaryClass={cs.IR},
      url={https://arxiv.org/abs/2312.14037}, 
}

@misc{xu2023tool,
      title={On the Tool Manipulation Capability of Open-source Large Language Models}, 
      author={Qiantong Xu and Fenglu Hong and Bo Li and Changran Hu and Zhengyu Chen and Jian Zhang},
      year={2023},
      eprint={2305.16504},
      archivePrefix={arXiv},
      primaryClass={cs.CL},
      url={https://arxiv.org/abs/2305.16504}, 
}

@misc{shen2024task,
      title={TaskBench: Benchmarking Large Language Models for Task Automation}, 
      author={Yongliang Shen and Kaitao Song and Xu Tan and Wenqi Zhang and Kan Ren and Siyu Yuan and Weiming Lu and Dongsheng Li and Yueting Zhuang},
      year={2024},
      eprint={2311.18760},
      archivePrefix={arXiv},
      primaryClass={cs.CL},
      url={https://arxiv.org/abs/2311.18760}, 
}

@misc{patil2023bfcl,
      title={Gorilla: Large Language Model Connected with Massive APIs}, 
      author={Shishir G. Patil and Tianjun Zhang and Xin Wang and Joseph E. Gonzalez},
      year={2023},
      eprint={2305.15334},
      archivePrefix={arXiv},
      primaryClass={cs.CL},
      url={https://arxiv.org/abs/2305.15334}, 
}

@misc{yang2025qwen3,
      title={Qwen3 Technical Report}, 
      author={An Yang and Anfeng Li and Baosong Yang and Beichen Zhang and Binyuan Hui and Bo Zheng and Bowen Yu and Chang Gao and Chengen Huang and Chenxu Lv and Chujie Zheng and Dayiheng Liu and Fan Zhou and Fei Huang and Feng Hu and Hao Ge and Haoran Wei and Huan Lin and Jialong Tang and Jian Yang and Jianhong Tu and Jianwei Zhang and Jianxin Yang and Jiaxi Yang and Jing Zhou and Jingren Zhou and Junyang Lin and Kai Dang and Keqin Bao and Kexin Yang and Le Yu and Lianghao Deng and Mei Li and Mingfeng Xue and Mingze Li and Pei Zhang and Peng Wang and Qin Zhu and Rui Men and Ruize Gao and Shixuan Liu and Shuang Luo and Tianhao Li and Tianyi Tang and Wenbiao Yin and Xingzhang Ren and Xinyu Wang and Xinyu Zhang and Xuancheng Ren and Yang Fan and Yang Su and Yichang Zhang and Yinger Zhang and Yu Wan and Yuqiong Liu and Zekun Wang and Zeyu Cui and Zhenru Zhang and Zhipeng Zhou and Zihan Qiu},
      year={2025},
      eprint={2505.09388},
      archivePrefix={arXiv},
      primaryClass={cs.CL},
      url={https://arxiv.org/abs/2505.09388}, 
}

@misc{jiang2023mistral7b,
      title={Mistral 7B}, 
      author={Albert Q. Jiang and Alexandre Sablayrolles and Arthur Mensch and Chris Bamford and Devendra Singh Chaplot and Diego de las Casas and Florian Bressand and Gianna Lengyel and Guillaume Lample and Lucile Saulnier and Lélio Renard Lavaud and Marie-Anne Lachaux and Pierre Stock and Teven Le Scao and Thibaut Lavril and Thomas Wang and Timothée Lacroix and William El Sayed},
      year={2023},
      eprint={2310.06825},
      archivePrefix={arXiv},
      primaryClass={cs.CL},
      url={https://arxiv.org/abs/2310.06825}, 
}

@misc{wei2023chain,
      title={Chain-of-Thought Prompting Elicits Reasoning in Large Language Models}, 
      author={Jason Wei and Xuezhi Wang and Dale Schuurmans and Maarten Bosma and Brian Ichter and Fei Xia and Ed Chi and Quoc Le and Denny Zhou},
      year={2023},
      eprint={2201.11903},
      archivePrefix={arXiv},
      primaryClass={cs.CL},
      url={https://arxiv.org/abs/2201.11903}, 
}

@article{10.1561/1500000019,
author = {Robertson, Stephen and Zaragoza, Hugo},
title = {The Probabilistic Relevance Framework: BM25 and Beyond},
year = {2009},
issue_date = {April 2009},
publisher = {Now Publishers Inc.},
address = {Hanover, MA, USA},
volume = {3},
number = {4},
issn = {1554-0669},
url = {https://doi.org/10.1561/1500000019},
doi = {10.1561/1500000019},
journal = {Found. Trends Inf. Retr.},
month = apr,
pages = {333–389},
numpages = {57}
}

@inproceedings{xie2025survey,
    title={A Survey on Personalized and Pluralistic Preference Alignment in Large Language Models},
    author={Zhouhang Xie and Junda Wu and Yiran Shen and Raghav Jain and Yu Xia and Xintong Li and Aaron Chang and Ryan A. Rossi and Tong Yu and Sachin Kumar and Bodhisattwa Prasad Majumder and Jingbo Shang and Prithviraj Ammanabrolu and Julian McAuley},
    booktitle={Second Conference on Language Modeling},
    year={2025},
    url={https://openreview.net/forum?id=lSWOMjonL7}
}

@inproceedings{yu2025explainablecot,
    title = "Explainable Chain-of-Thought Reasoning: An Empirical Analysis on State-Aware Reasoning Dynamics",
    author = "Yu, Sheldon  and
      Xiong, Yuxin  and
      Wu, Junda  and
      Li, Xintong  and
      Yu, Tong  and
      Chen, Xiang  and
      Sinha, Ritwik  and
      Shang, Jingbo  and
      McAuley, Julian",
    editor = "Christodoulopoulos, Christos  and
      Chakraborty, Tanmoy  and
      Rose, Carolyn  and
      Peng, Violet",
    booktitle = "Findings of the Association for Computational Linguistics: EMNLP 2025",
    month = nov,
    year = "2025",
    address = "Suzhou, China",
    publisher = "Association for Computational Linguistics",
    url = "https://aclanthology.org/2025.findings-emnlp.904/",
    doi = "10.18653/v1/2025.findings-emnlp.904",
    pages = "16660--16667",
    ISBN = "979-8-89176-335-7",
}

@inproceedings{wu2025ctrls,
  title={CTRLS: Chain-of-Thought Reasoning via Latent State-Transition},
  author={Wu, Junda and Xiong, Yuxin and Li, Xintong and Yu, Sheldon and Hu, Zhengmian and Yu, Tong and Wang, Rui and Chen, Xiang and Shang, Jingbo and McAuley, Julian},
  booktitle={The 29th International Conference on Artificial Intelligence and Statistics}
}

@inproceedings{wu2024decot,
  title={Decot: Debiasing chain-of-thought for knowledge-intensive tasks in large language models via causal intervention},
  author={Wu, Junda and Yu, Tong and Chen, Xiang and Wang, Haoliang and Rossi, Ryan and Kim, Sungchul and Rao, Anup and McAuley, Julian},
  booktitle={Proceedings of the 62nd Annual Meeting of the Association for Computational Linguistics (Volume 1: Long Papers)},
  pages={14073--14087},
  year={2024}
}

@inproceedings{wu2025ocean,
  title={Ocean: Offline chain-of-thought evaluation and alignment in large language models},
  author={Wu, Junda and Li, Xintong and Wang, Ruoyu and Xia, Yu and Xiong, Yuxin and Wang, Jianing and Yu, Tong and Chen, Xiang and Kveton, Branislav and Yao, Lina and others},
  booktitle={International Conference on Learning Representations},
  volume={2025},
  pages={100570--100589},
  year={2025}
}

@inproceedings{wu2025docreact,
  title={Doc-react: Multi-page heterogeneous document question-answering},
  author={Wu, Junda and Xia, Yu and Yu, Tong and Chen, Xiang and Harsha, Sai Sree and Maharaj, Akash V and Zhang, Ruiyi and Bursztyn, Victor and Kim, Sungchul and Rossi, Ryan A and others},
  booktitle={Proceedings of the 63rd Annual Meeting of the Association for Computational Linguistics (Volume 2: Short Papers)},
  pages={67--78},
  year={2025}
}

@inproceedings{wu2021clusterbandit,
  title={Clustering of conversational bandits for user preference learning and elicitation},
  author={Wu, Junda and Zhao, Canzhe and Yu, Tong and Li, Jingyang and Li, Shuai},
  booktitle={Proceedings of the 30th ACM International Conference on Information \& Knowledge Management},
  pages={2129--2139},
  year={2021}
}

@article{huang2025pluralistic,
  title={Pluralistic Off-policy Evaluation and Alignment},
  author={Huang, Chengkai and Wu, Junda and Xie, Zhouhang and Xia, Yu and Wang, Rui and Yu, Tong and Mitra, Subrata and McAuley, Julian and Yao, Lina},
  journal={arXiv preprint arXiv:2509.19333},
  year={2025}
}

@inproceedings{wu2022dynamicsrl,
  title={Dynamics-aware adaptation for reinforcement learning based cross-domain interactive recommendation},
  author={Wu, Junda and Xie, Zhihui and Yu, Tong and Zhao, Handong and Zhang, Ruiyi and Li, Shuai},
  booktitle={Proceedings of the 45th international ACM SIGIR conference on research and development in information retrieval},
  pages={290--300},
  year={2022}
}

@inproceedings{hu2025hiersucb,
    title={Interactive Visualization Recommendation with Hier-{SUCB}},
    author={Songwen Hu and Ryan A. Rossi and Tong Yu and Junda Wu and Handong Zhao and Sungchul Kim and Shuai Li},
    booktitle={THE WEB CONFERENCE 2025},
    year={2025},
    url={https://openreview.net/forum?id=Hkh2umURYm}
}

@inproceedings{xia2023deconfoundedcrs,
  title={User-regulation deconfounded conversational recommender system with bandit feedback},
  author={Xia, Yu and Wu, Junda and Yu, Tong and Kim, Sungchul and Rossi, Ryan A and Li, Shuai},
  booktitle={Proceedings of the 29th ACM SIGKDD conference on knowledge discovery and data mining},
  pages={2694--2704},
  year={2023}
}

@inproceedings{xia2025sand,
  title={SAND: Boosting LLM agents with self-taught action deliberation},
  author={Xia, Yu and Shen, Yiran Jenny and Wu, Junda and Yu, Tong and Kim, Sungchul and Rossi, Ryan A and Yao, Lina and McAuley, Julian},
  booktitle={Proceedings of the 2025 Conference on Empirical Methods in Natural Language Processing},
  pages={3062--3077},
  year={2025}
}

@article{wang2025dice,
  title={Dice: Dynamic in-context example selection in llm agents via efficient knowledge transfer},
  author={Wang, Ruoyu and Wu, Junda and Xia, Yu and Yu, Tong and Rossi, Ryan A and McAuley, Julian and Yao, Lina},
  journal={arXiv preprint arXiv:2507.23554},
  year={2025}
}

@article{kveton2025aldpo,
  title={Active learning for direct preference optimization},
  author={Kveton, Branislav and Li, Xintong and McAuley, Julian and Rossi, Ryan and Shang, Jingbo and Wu, Junda and Yu, Tong},
  journal={arXiv preprint arXiv:2503.01076},
  year={2025}
}

@inproceedings{wu2023infoprompt,
    title={InfoPrompt: Information-Theoretic Soft Prompt Tuning for Natural Language Understanding},
    author={Junda Wu and Tong Yu and Rui Wang and Zhao Song and Ruiyi Zhang and Handong Zhao and Chaochao Lu and Shuai Li and Ricardo Henao},
    booktitle={Thirty-seventh Conference on Neural Information Processing Systems},
    year={2023},
    url={https://openreview.net/forum?id=mSNfjOcDUv}
}

@article{wu2024mllmsurvey,
  title={Personalized multimodal large language models: A survey},
  author={Wu, Junda and Lyu, Hanjia and Xia, Yu and Zhang, Zhehao and Barrow, Joe and Kumar, Ishita and Mirtaheri, Mehrnoosh and Chen, Hongjie and Rossi, Ryan A and Dernoncourt, Franck and others},
  journal={arXiv preprint arXiv:2412.02142},
  year={2024}
}

@article{zhang2024llmpersonalization,
  title={Personalization of large language models: A survey},
  author={Zhang, Zhehao and Rossi, Ryan A and Kveton, Branislav and Shao, Yijia and Yang, Diyi and Zamani, Hamed and Dernoncourt, Franck and Barrow, Joe and Yu, Tong and Kim, Sungchul and others},
  journal={arXiv preprint arXiv:2411.00027},
  year={2024}
}

@misc{surana2026generatefiltercontrolreplay,
      title={Generate, Filter, Control, Replay: A Comprehensive Survey of Rollout Strategies for LLM Reinforcement Learning}, 
      author={Rohan Surana and Gagan Mundada and Xunyi Jiang and Chuhan Wang and Zhenwei Tang and Difan Jiao and Zihan Huang and Yuxin Xiong and Junda Wu and Sheldon Yu and Xintong Li and Raghav Jain and Nikki Kuang and Sizhe Zhou and Bowen Jin and Zhendong Chu and Tong Yu and Ryan Rossi and Kuan-Hao Huang and Jingbo Shang and Jiawei Han and Julian McAuley},
      year={2026},
      eprint={2605.02913},
      archivePrefix={arXiv},
      primaryClass={cs.LG},
      url={https://arxiv.org/abs/2605.02913}, 
}

@article{wu2024visualpromptsurvey,
  title={Visual prompting in multimodal large language models: A survey},
  author={Wu, Junda and Zhang, Zhehao and Xia, Yu and Li, Xintong and Xia, Zhaoyang and Chang, Aaron and Yu, Tong and Kim, Sungchul and Rossi, Ryan A and Zhang, Ruiyi and others},
  journal={arXiv preprint arXiv:2409.15310},
  year={2024}
}

@inproceedings{nguyen2025guisurvey,
  title={Gui agents: A survey},
  author={Nguyen, Dang and Chen, Jian and Wang, Yu and Wu, Gang and Park, Namyong and Hu, Zhengmian and Lyu, Hanjia and Wu, Junda and Aponte, Ryan and Xia, Yu and others},
  booktitle={Findings of the Association for Computational Linguistics: ACL 2025},
  pages={22522--22538},
  year={2025}
}

@inproceedings{wu2025icrpo,
  title={In-context Ranking Preference Optimization},
  author={Wu, Junda and Surana, Rohan and Xie, Zhouhang and Shen, Yiran and Xia, Yu and Yu, Tong and Rossi, Ryan A and Ammanabrolu, Prithviraj and McAuley, Julian},
  booktitle={Second Conference on Language Modeling}
}

@inproceedings{wang2025selfupdatable,
  title={Self-updatable large language models by integrating context into model parameters},
  author={Wang, Yu and Liu, Xinshuang and Chen, Xiusi and OBrien, Sean and Wu, Junda and McAuley, Julian},
  booktitle={International Conference on Learning Representations},
  volume={2025},
  pages={16961--16979},
  year={2025}
}

@article{mundada2026wsgrpo,
  title={WS-GRPO: Weakly-Supervised Group-Relative Policy Optimization for Rollout-Efficient Reasoning},
  author={Mundada, Gagan and Huang, Zihan and Surana, Rohan and Yu, Sheldon and Zhang, Jennifer Yuntong and Li, Xintong and Yu, Tong and Yao, Lina and Shang, Jingbo and McAuley, Julian and others},
  journal={arXiv preprint arXiv:2602.17025},
  year={2026}
}

@misc{huang2025agenticrs,
      title={Towards Agentic Recommender Systems in the Era of Multimodal Large Language Models}, 
      author={Chengkai Huang and Junda Wu and Yu Xia and Zixu Yu and Ruhan Wang and Tong Yu and Ruiyi Zhang and Ryan A. Rossi and Branislav Kveton and Dongruo Zhou and Julian McAuley and Lina Yao},
      year={2025},
      eprint={2503.16734},
      archivePrefix={arXiv},
      primaryClass={cs.AI},
      url={https://arxiv.org/abs/2503.16734}, 
}

@misc{lu2024uci,
      title={Improving Multi-candidate Speculative Decoding}, 
      author={Xiaofan Lu and Yixiao Zeng and Feiyang Ma and Zixu Yu and Marco Levorato},
      year={2024},
      eprint={2409.10644},
      archivePrefix={arXiv},
      primaryClass={cs.CL},
      url={https://arxiv.org/abs/2409.10644}, 
}
\bibliographystyle{colm2026_conference}

\appendix
\newpage
\section{Appendix}

\subsection{Implementation Algorithm}
\label{sec:method_algo}

Algorithm~\ref{alg:olivia} summarizes the complete \textsc{OLIVIA} procedure.
Each episode begins by receiving a task and initializing the valid action set.
At every decision step, the frozen LLM generates a reasoning trace and
produces the hidden-state context (line~6), the controller scores all valid
actions via LinUCB (line~8) and selects the maximizer (line~10). After
observing the step-level binary reward, the selected action's sufficient
statistics are updated in place using the Sherman--Morrison rank-one formula
(lines~16--17), requiring no gradient computation or backward pass. The
per-action inverse covariance and reward accumulator persist across episodes,
enabling the controller to improve over the full deployment stream.

\begin{algorithm}[htb]
\caption{\textsc{OLIVIA}: Online Contextual Bandit for ReAct Tool Selection}
\label{alg:olivia}
\begin{algorithmic}[1]
\STATE \textbf{Inputs}: exploration parameter $\alpha>0$, extra steps $s$, max selections per arm $m$
\STATE \textbf{Initialize}: For each arm $a\!\in\!\mathcal{A}$:\; $A_a^{-1} \gets I_d$,\; $b_a \gets \mathrm{emb}_{\mathrm{LLM}}(a)$,\; $n_a \gets 0$
\FOR{episode $e=1$ to $T$}
    \STATE Receive task $q_e$ with ground-truth tool multiset $G_e$;\; set $\mathcal{V} \gets \mathcal{A}$,\; $L_e \gets |G_e|+s$
    \FOR{$t=1$ to $L_e$}
        \STATE Generate thought $z_{e,t}$ via LLM conditioned on $(q_e,\, z_{e,<t},\, a_{e,<t})$
        \STATE Extract context $x_{e,t} \gets h_{\mathrm{LLM}}\!\bigl(q_e,\, z_{e,1:t},\, a_{e,<t}\bigr)$ \hfill\COMMENT{last hidden state at action position}
        \FORALL{$a \in \mathcal{V}$}
            \STATE $\theta_a \gets A_a^{-1} b_a$,\quad $\mathrm{ucb}_{e,t}(a) \gets \theta_a^\top x_{e,t} \;+\; \alpha\,\sqrt{x_{e,t}^\top A_a^{-1} x_{e,t}}$
        \ENDFOR
        \STATE Select $a_{e,t} \gets \arg\max_{a \in \mathcal{V}}\;\mathrm{ucb}_{e,t}(a)$
        \STATE Execute $a_{e,t}$;\; observe per-step reward $r_{e,t} = \mathbf{1}\!\bigl[a_{e,t} \in G_e^{\mathrm{rem}}\bigr]$
        \IF{$r_{e,t}=1$}
            \STATE Decrement remaining count $G_e^{\mathrm{rem}}(a_{e,t})$
        \ENDIF
        \STATE Update selection count $n_{a_{e,t}} \mathrel{+}= 1$;\; if $n_{a_{e,t}} \ge m$ remove $a_{e,t}$ from $\mathcal{V}$
        \STATE \textbf{Policy update} (Sherman--Morrison):
        \STATE \quad $k \gets A_{a_{e,t}}^{-1} x_{e,t}$
        \STATE \quad $A_{a_{e,t}}^{-1} \gets A_{a_{e,t}}^{-1} - \dfrac{k\,k^\top}{1 + x_{e,t}^\top k}$,\qquad $b_{a_{e,t}} \gets b_{a_{e,t}} + r_{e,t}\, x_{e,t}$
        \IF{$\mathcal{V} = \emptyset$}
            \STATE \textbf{break} \COMMENT{all arms exhausted}
        \ENDIF
    \ENDFOR
    \STATE Compute episode metrics: $\mathrm{recall}_e,\, \mathrm{precision}_e,\, \mathrm{F1}_e$
\ENDFOR
\end{algorithmic}
\end{algorithm}

\subsection{Case Study}
\label{sec:case_study}

We present qualitative examples from the ToolBench2 benchmark
(\texttt{Qwen3-4B}, $\alpha{=}0.1$) to illustrate how \textsc{OLIVIA}'s
online controller improves tool selection over the episode stream.

\paragraph{Learning progression.}
The following two boxes show the \emph{same task} encountered early
(episode~3) and late (episode~935) in the stream. Early on, the controller
has not yet accumulated feedback and selects entirely unrelated tools
(F1\,=\,0.00). By episode~935, online updates have shifted the per-action
estimates so that all three ground-truth tools are recovered (recall\,=\,1.00),
despite the first three actions still being misses---demonstrating that the
UCB exploration bonus eventually directs selection toward the correct tools.

\begin{tcolorbox}[colback=gray!5,colframe=gray!75,title=Episode 3 --- Early Stream (F1\,=\,0.00)]
\small
\textbf{Task:} I'm organizing a team building activity for my company. Can you help me create a short URL for the event details? Also, suggest some domain options for the team building link. Furthermore, provide me with a list of available domains for branding the short link.

\textbf{Ground truth (3):} bitly.shorten, URL Link Shortener.Create a new link, URL Link Shortener.Get a list of domains

\begin{enumerate}[leftmargin=*,itemsep=1pt]
  \item \textit{Thought:} The user needs a short URL for the event details.\\
        \textbf{Action:} OTT details.New Arrivals \textcolor{red!70!black}{\ding{55}}
  \item \textit{Thought:} The user also wants domain options for the short link.\\
        \textbf{Action:} Streaming Availability.Get Basic \textcolor{red!70!black}{\ding{55}}
  \item \textit{Thought:} Additionally, the user needs a list of available domains for branding.\\
        \textbf{Action:} Similar Movies.find similar \textcolor{red!70!black}{\ding{55}}
  \item \textit{Thought:} Now, I need to check if all the actions have been completed.\\
        \textbf{Action:} Watchmode.Streaming Release Dates \textcolor{red!70!black}{\ding{55}}
  \item \textit{Thought:} All required actions have been completed.\\
        \textbf{Action:} Vimeo.GetRelatedChannels \textcolor{red!70!black}{\ding{55}}
  \item \textit{Thought:} The user's task is complete.\\
        \textbf{Action:} Movie, TV, music search and download.Get Monthly Top 100 Music Torrents \textcolor{red!70!black}{\ding{55}}
\end{enumerate}

\textbf{Result:} P=0.00, R=0.00, F1=0.00
\end{tcolorbox}

\begin{tcolorbox}[colback=green!3,colframe=green!60!black,title=Episode 935 --- Late Stream (F1\,=\,0.67)]
\small
\textbf{Task:} I'm organizing a team building activity for my company. Can you help me create a short URL for the event details? Also, suggest some domain options for the team building link. Furthermore, provide me with a list of available domains for branding the short link.

\textbf{Ground truth (3):} bitly.shorten, URL Link Shortener.Create a new link, URL Link Shortener.Get a list of domains

\begin{enumerate}[leftmargin=*,itemsep=1pt]
  \item \textit{Thought:} The user needs a short URL for the event details.\\
        \textbf{Action:} Watchmode.Title Seasons \textcolor{red!70!black}{\ding{55}}
  \item \textit{Thought:} The user also wants domain options for the short link.\\
        \textbf{Action:} Watchmode.New Titles \textcolor{red!70!black}{\ding{55}}
  \item \textit{Thought:} Additionally, the user is looking for a list of available domains for branding.\\
        \textbf{Action:} Streaming Availability.Search Pro \textcolor{red!70!black}{\ding{55}}
  \item \textit{Thought:} Now, I need to check if all the actions have been completed.\\
        \textbf{Action:} URL Link Shortener.Create a new link \textcolor{green!70!black}{\ding{51}}
  \item \textit{Thought:} All tasks are completed.\\
        \textbf{Action:} bitly.shorten \textcolor{green!70!black}{\ding{51}}
  \item \textit{Thought:} The final answer is ready.\\
        \textbf{Action:} URL Link Shortener.Get a list of domains \textcolor{green!70!black}{\ding{51}}
\end{enumerate}

\textbf{Result:} P=0.50, R=1.00, F1=0.67
\end{tcolorbox}

\paragraph{Learned controller.}
By episode~1130, the controller has accumulated enough feedback to select all
three ground-truth tools within the first four steps, even on a previously
unseen task phrasing.

\begin{tcolorbox}[colback=green!3,colframe=green!60!black,title=Episode 1130 --- Complex Task{,} 7/9 GT Tools (F1\,=\,0.67)]
\small
\textbf{Task:} I'm organizing a movie marathon for my company's team-building event. Could you suggest some action movies from the 90s that are available for streaming on Netflix, Prime Video, and Hulu? Please include the IMDb ratings and streaming links for each movie.

\textbf{Ground truth (9):} Watchmode.List Titles, Watchmode.New People, Watchmode.Title Seasons, OTT details.Basic Info, OTT details.Advanced Search, OTT details.Title Details, Streaming Availability.Search By Title, Streaming Availability.Search Ultra, Streaming Availability.Search Pro

\begin{enumerate}[leftmargin=*,itemsep=1pt]
  \item \textit{Thought:} The user is looking for action movies from the 90s.\\
        \textbf{Action:} Vimeo.SearchVideos \textcolor{red!70!black}{\ding{55}}
  \item \textit{Thought:} The user wants to know where these movies are available for streaming.\\
        \textbf{Action:} Watchmode.List Titles \textcolor{green!70!black}{\ding{51}}
  \item \textit{Thought:} The user is interested in IMDb ratings for these movies.\\
        \textbf{Action:} Streaming Availability.Search Ultra \textcolor{green!70!black}{\ding{51}}
  \item \textit{Thought:} The user wants the actual streaming links for each movie.\\
        \textbf{Action:} OTT details.Basic Info \textcolor{green!70!black}{\ding{51}}
  \item \textit{Thought:} The information should be accurate and up-to-date for a company event.\\
        \textbf{Action:} Streaming Availability.Search Pro \textcolor{green!70!black}{\ding{51}}
  \item \textit{Thought:} The information needs to be concise and easy to share with the team.\\
        \textbf{Action:} Watchmode.Title Seasons \textcolor{green!70!black}{\ding{51}}
  \item \textit{Thought:} The user is likely looking for a variety of options to cater to different tastes.\\
        \textbf{Action:} OTT details.Title Details \textcolor{green!70!black}{\ding{51}}
  \item \textit{Thought:} The user might also want to know the release years to ensure they are from the 90s.\\
        \textbf{Action:} OTT details.Search \textcolor{red!70!black}{\ding{55}}
  \item \textit{Thought:} The information should be organized in a way that is easy to use during the marathon.\\
        \textbf{Action:} OTT details.Additional Title Details \textcolor{red!70!black}{\ding{55}}
  \item \textit{Thought:} The user is looking for movies that can be watched in a short time frame.\\
        \textbf{Action:} Streaming Availability.Search By Title \textcolor{green!70!black}{\ding{51}}
  \item \textit{Thought:} The movies should be engaging and suitable for a group setting.\\
        \textbf{Action:} Similar Movies.find similar \textcolor{red!70!black}{\ding{55}}
  \item \textit{Thought:} The user is looking for movies that are well-received with good action elements.\\
        \textbf{Action:} Vimeo.GetChannelVideos \textcolor{red!70!black}{\ding{55}}
\end{enumerate}

\textbf{Result:} P=0.58, R=0.78, F1=0.67
\end{tcolorbox}

\subsection{Per-Benchmark Learning Curves}
\label{sec:app_curves}

We report per-benchmark learning curves for both backbone models. Each plot
shows running-average F1 over the episode stream. Non-learning baselines
(CoT, BM25, ReAct) appear as flat reference lines; online methods (Bandit,
CLIN, \textsc{OLIVIA}) evolve as they accumulate feedback.

\paragraph{Qwen3-4B.}

\begin{figure}[h]
\centering
\begin{minipage}[t]{0.48\textwidth}
  \centering
  \includegraphics[width=\linewidth]{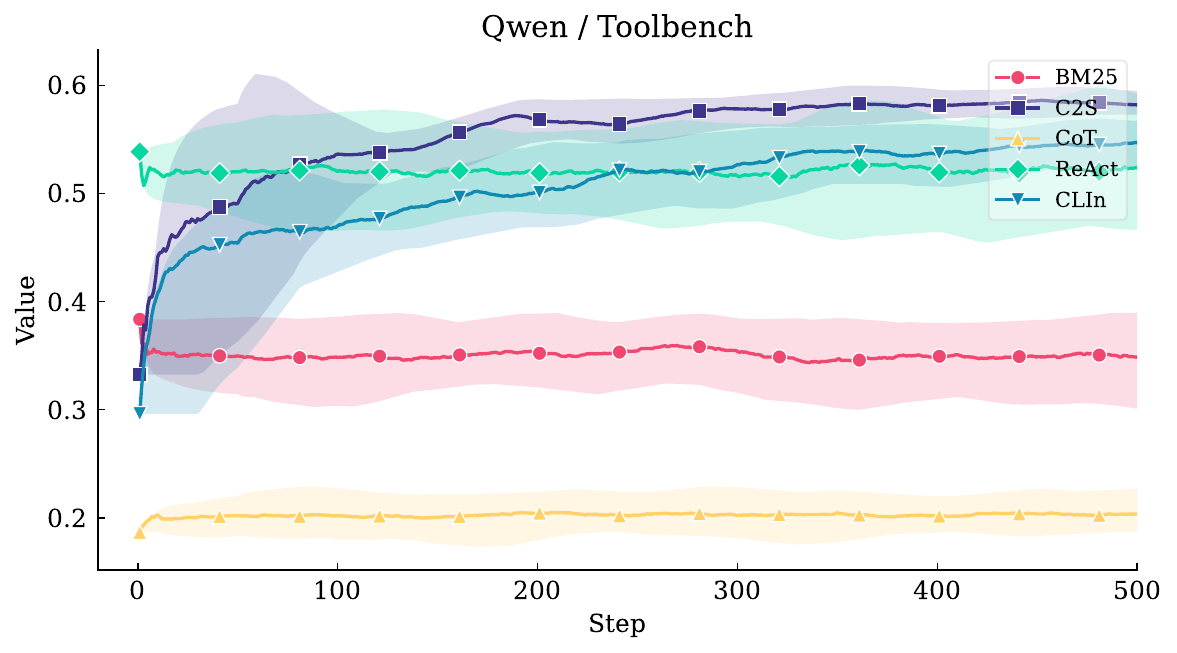}
  \captionof{figure}{Qwen3-4B on ToolBench.}
  \label{fig:app_qwen_toolbench}
\end{minipage}\hfill
\begin{minipage}[t]{0.48\textwidth}
  \centering
  \includegraphics[width=\linewidth]{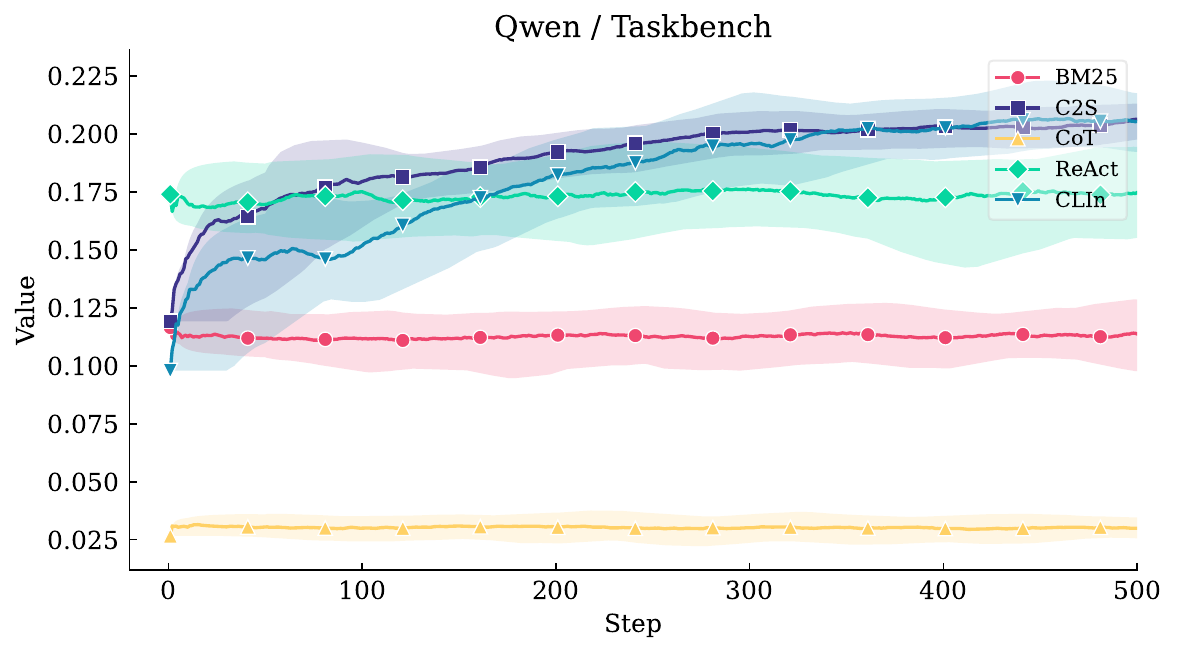}
  \captionof{figure}{Qwen3-4B on TaskBench.}
  \label{fig:app_qwen_taskbench}
\end{minipage}

\vspace{0.5em}

\begin{minipage}[t]{0.48\textwidth}
  \centering
  \includegraphics[width=\linewidth]{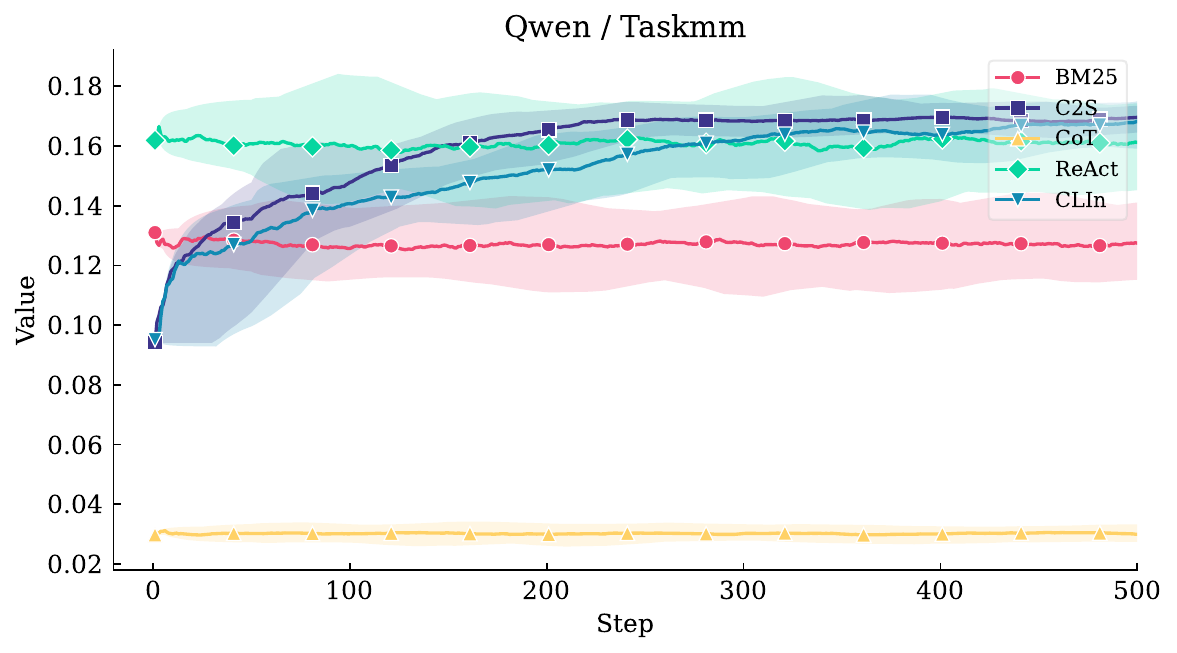}
  \captionof{figure}{Qwen3-4B on TaskBench-MM.}
  \label{fig:app_qwen_taskmm}
\end{minipage}\hfill
\begin{minipage}[t]{0.48\textwidth}
  \centering
  \includegraphics[width=\linewidth]{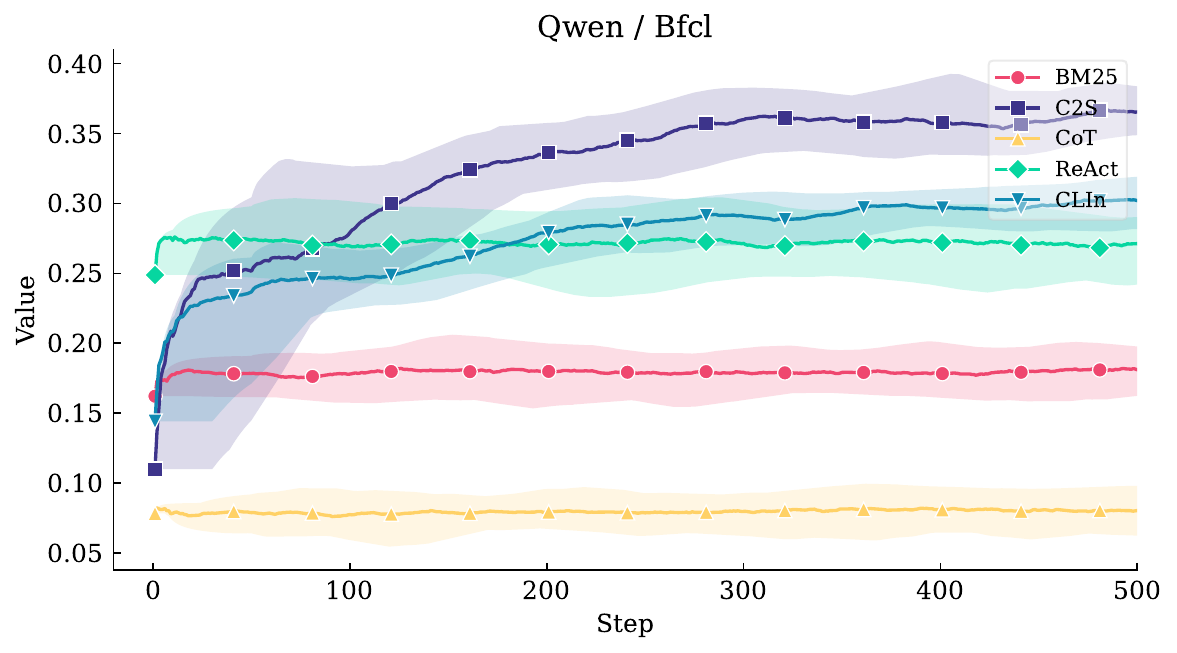}
  \captionof{figure}{Qwen3-4B on BFCL.}
  \label{fig:app_qwen_bfcl}
\end{minipage}
\end{figure}

\paragraph{Mistral-7B.}

\begin{figure}[h]
\centering
\begin{minipage}[t]{0.48\textwidth}
  \centering
  \includegraphics[width=\linewidth]{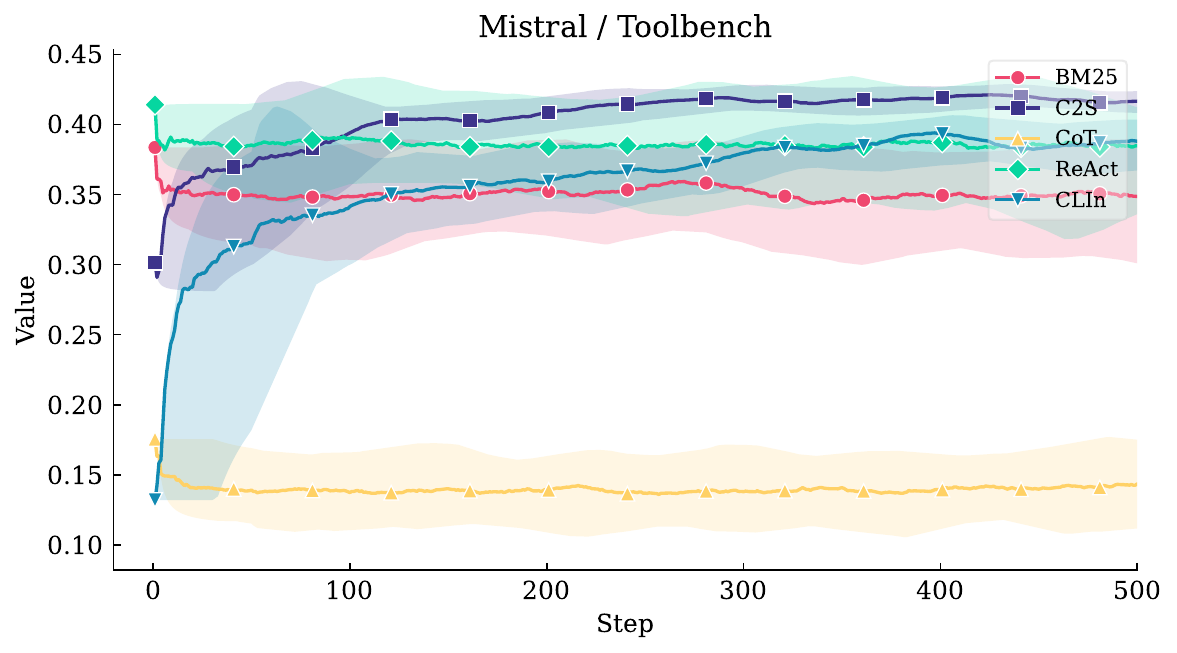}
  \captionof{figure}{Mistral-7B on ToolBench.}
  \label{fig:app_mistral_toolbench}
\end{minipage}\hfill
\begin{minipage}[t]{0.48\textwidth}
  \centering
  \includegraphics[width=\linewidth]{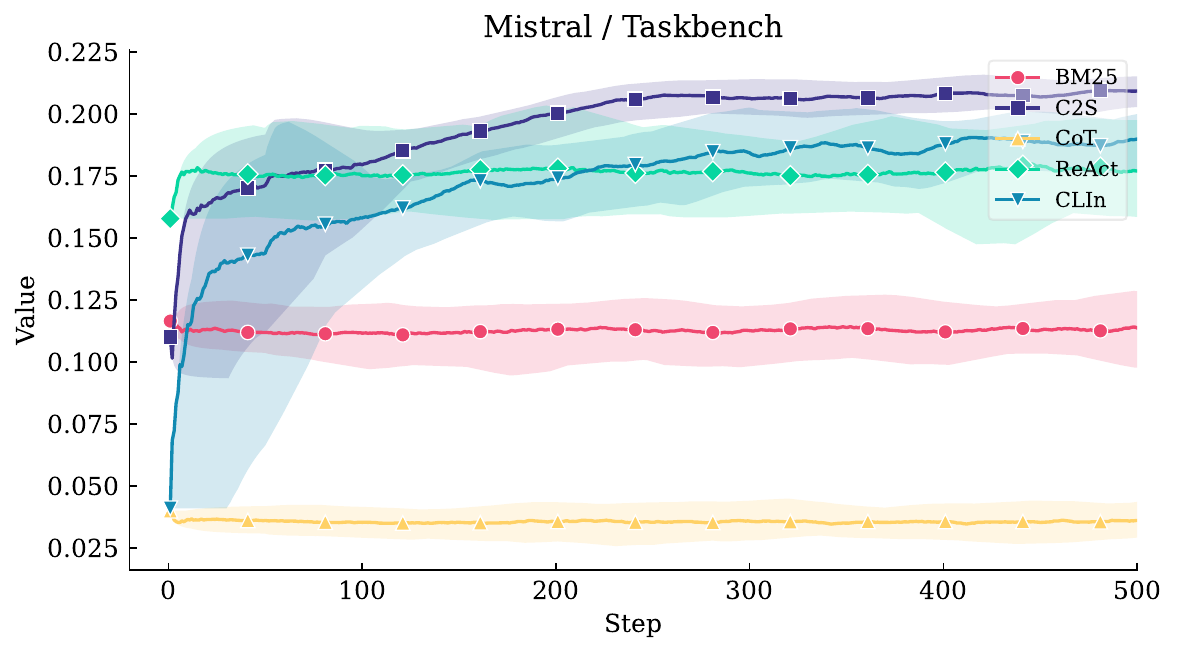}
  \captionof{figure}{Mistral-7B on TaskBench.}
  \label{fig:app_mistral_taskbench}
\end{minipage}

\vspace{0.5em}

\begin{minipage}[t]{0.48\textwidth}
  \centering
  \includegraphics[width=\linewidth]{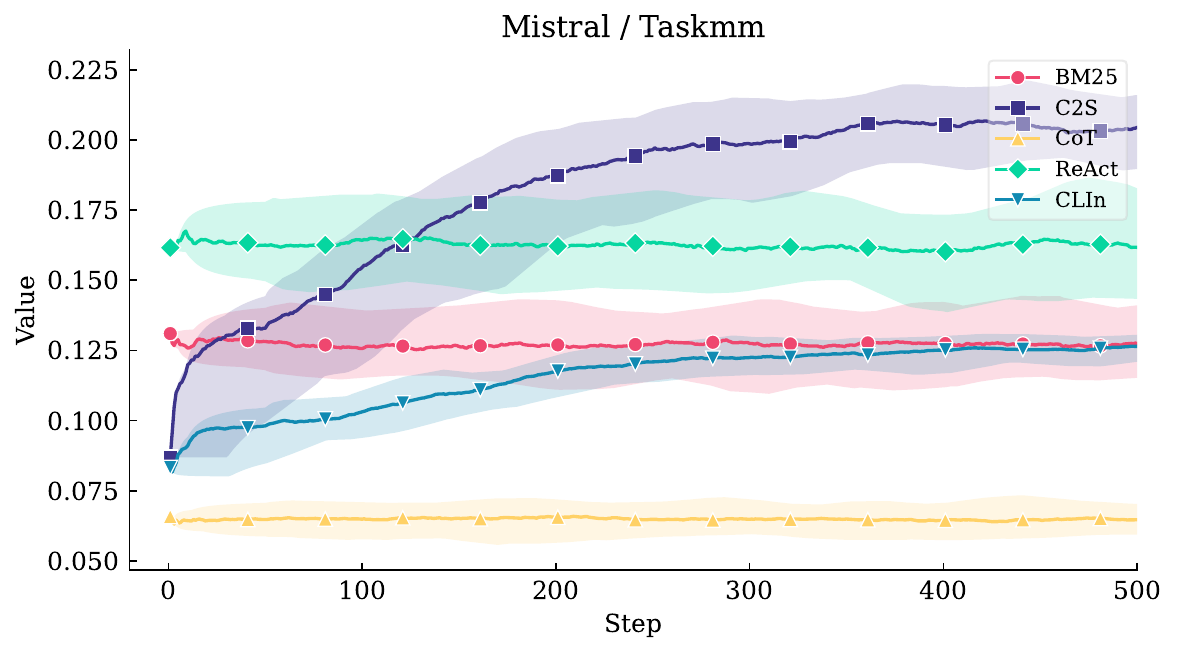}
  \captionof{figure}{Mistral-7B on TaskBench-MM.}
  \label{fig:app_mistral_taskmm}
\end{minipage}\hfill
\begin{minipage}[t]{0.48\textwidth}
  \centering
  \includegraphics[width=\linewidth]{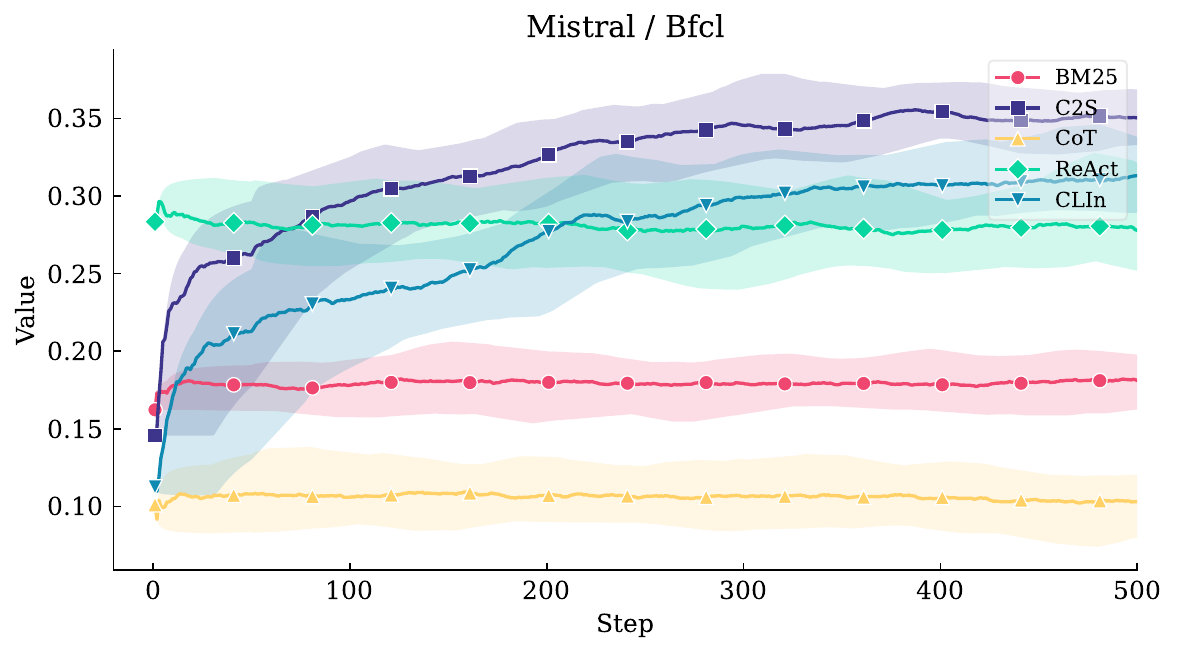}
  \captionof{figure}{Mistral-7B on BFCL.}
  \label{fig:app_mistral_bfcl}
\end{minipage}
\end{figure}

\subsection{Reward Studies}
Each panel shows the three reward metrics (Avg Recall, Avg F1, Avg Precision) with the same smoothing, bands, and styling as before. The shared legend sits at the top center, and only the bottom row has x-axis labels / left column has y-axis labels to keep it clean.

\begin{figure}
\begin{minipage}[t]{1\textwidth}
  \centering
  \includegraphics[width=\linewidth]{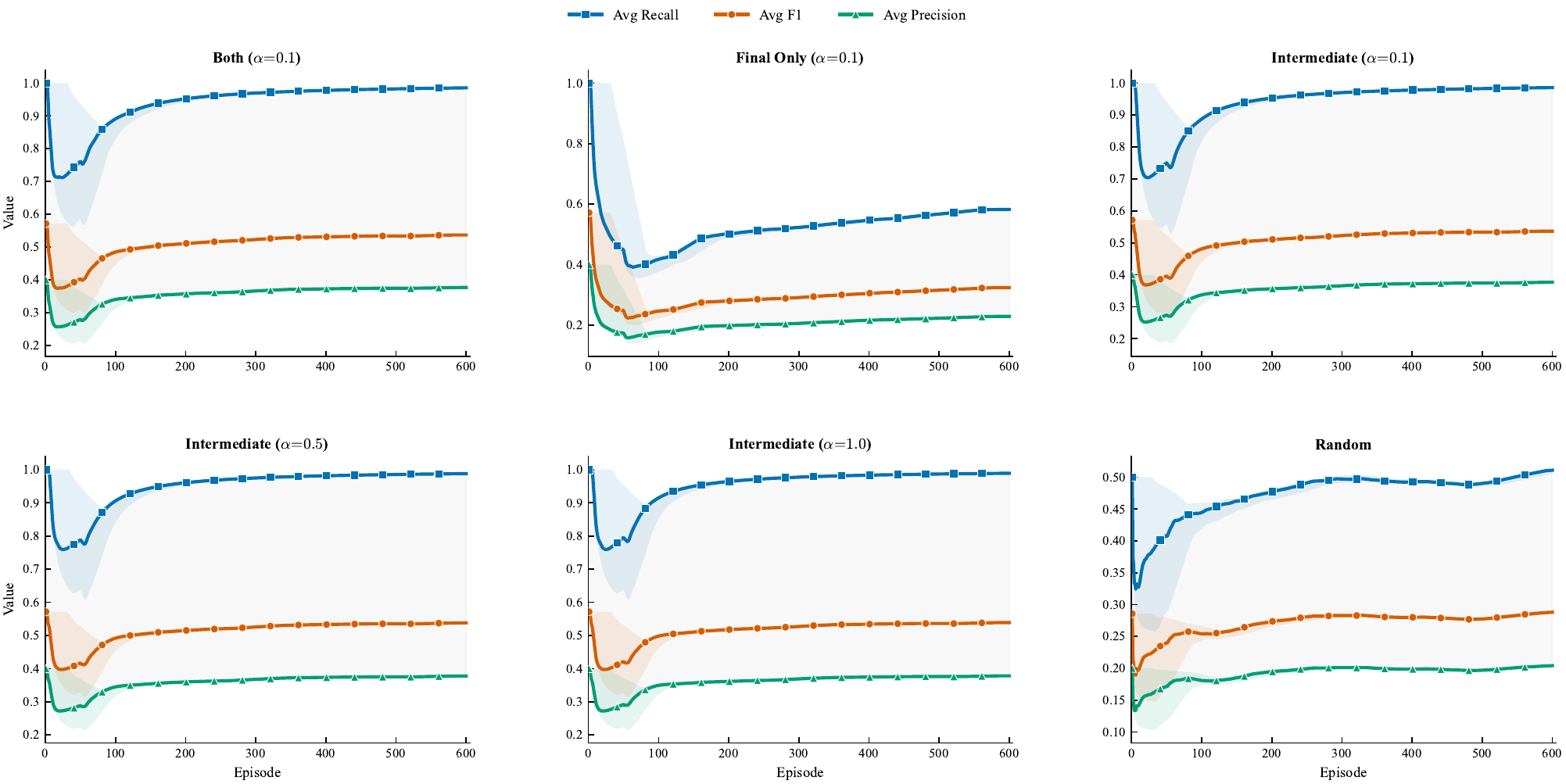}
  \captionof{figure}{Reward Studies.}
  \label{fig:app_mistral_bfcl}
\end{minipage}
\end{figure}

\end{document}